\documentclass[11pt]{article}

\usepackage[final]{acl}

\usepackage{times}
\usepackage{latexsym}
\usepackage[T1]{fontenc}
\usepackage[utf8]{inputenc}
\usepackage{microtype}
\usepackage{inconsolata}
\usepackage{graphicx}
\usepackage{booktabs}
\usepackage{multirow}
\usepackage{url}
\usepackage{float}
\usepackage{dblfloatfix}
\usepackage{hyperref}

\title{MKJ at SemEval-2026 Task 9: A Comparative Study of Generalist, Specialist, and Ensemble Strategies for Multilingual Polarization}

\author{Maziar Kianimoghadam Jouneghani \\
University of Turin \\
\texttt{maziar.kianimoghadam@edu.unito.it}}

\begin{document}
\maketitle

\begin{abstract}
We present a systematic study of multilingual polarization detection across 22 languages for SemEval-2026 Task 9 (Subtask 1), contrasting multilingual generalists with language-specific specialists and hybrid ensembles. While a standard generalist like XLM-RoBERTa suffices when its tokenizer aligns with the target text, it may struggle with distinct scripts (e.g., Khmer, Odia) where monolingual specialists yield significant gains. Rather than enforcing a single universal architecture, we adopt a language-adaptive selection strategy that chooses among multilingual generalists, language-specific specialists, and hybrid ensembles based on development performance. Additionally, cross-lingual augmentation via NLLB-200 yielded mixed results, often underperforming native architecture selection and degrading morphologically rich tracks. Our final system achieves an overall macro-averaged F1 score of \textbf{0.796} and an average accuracy of \textbf{0.826} across all 22 tracks. Code and final test predictions are publicly available at: \url{https://github.com/Maziarkiani/SemEval2026-Task9-Subtask1-Polarization}.
\end{abstract}

\section{Introduction}
Polarization detection---identifying rhetoric that reinforces ideological division---transcends simple sentiment analysis, requiring models to parse complex socio-political nuances. SemEval-2026 Task 9 \citep{naseem-etal-2026-polar} introduces the POLAR benchmark \citep{naseem2026polar} for this task across 22 typologically diverse languages. In this paper, we address \textbf{Subtask 1 only}, focusing on multilingual polarization detection across the 22 language tracks of the shared task. The central challenge is an architectural dilemma: balancing the semantic reasoning of massive multilingual generalists against the culturally grounded vocabularies of monolingual specialists.

In this work, we address three central research questions: (\textbf{RQ1}) Does XLM-RoBERTa provide a sufficient universal baseline across 22 diverse languages? (\textbf{RQ2}) When the baseline fails, under what linguistic conditions do monolingual specialists or alternative high-capacity generalists (e.g., mDeBERTa-v3) outperform it? (\textbf{RQ3}) Can cross-lingual data augmentation (via translation) compensate for resource scarcity?

To navigate these questions, we developed a language-adaptive selection strategy. Rather than forcing a universal architecture, we established a systematic empirical policy for architecture selection (detailed in Section 3). Table \ref{tab:model_strategy} details our final system configurations, highlighting where generalization succeeds and where specialization is mandatory.

\section{Related Work}
Polarization detection extends tasks like sentiment and stance analysis by requiring models to identify antagonistic rhetoric across diverse languages, scripts, and contexts. The POLAR benchmark and SemEval-2026 Task 9 formalize this challenge across 22 languages \citep{naseem-etal-2026-polar,naseem2026polar}.

While multilingual encoders like XLM-R \citep{conneau-etal-2020-unsupervised} are standard baselines, downstream performance is not dictated by scale alone. \citet{rust-etal-2021-how-good-tokenizer} show that monolingual tokenizers often outperform multilingual ones due to varied tokenization quality across languages. This directly motivates our fragmentation analysis and comparison of generalists versus language-specific specialists.

Ensembling is also a recurring strategy in competitive NLP, especially when different models exhibit complementary error profiles. The SemEval-2026 Task 9 overview reports that ensemble prediction, threshold tuning, and weighted fusion were among the most common strategies adopted by participating systems, particularly in strong submissions \citep{naseem-etal-2026-polar}. Similar ensemble-based approaches have also been used in recent SemEval systems for multilingual classification, where combining diverse models improves robustness across heterogeneous inputs \citep{zhu-2024-rdproj}.

Finally, while data augmentation addresses low-resource constraints, its efficacy is highly task-dependent. Several participating teams experimented with translation and paraphrasing \citep{naseem-etal-2026-polar}. However, recent work suggests synthetic data quality is highly method-dependent \citep{pecher-etal-2026-better-generators}. Consequently, our translation ablation is positioned not as a rejection of augmentation in general, but as evidence that naive forward translation was less reliable than architecture selection in our setting.

\begin{table*}[t!]
\centering
\small
\begin{tabular}{l l l l l}
\toprule
\textbf{Lang} & \textbf{Code} & \textbf{Strategy} & \textbf{Model Details} & \textbf{Weights \& $\tau$} \\
\midrule
\multicolumn{5}{c}{\textit{Group A: Generalist Sufficiency}} \\
Burmese & mya & Generalist & mDeBERTa-v3 (Base) \citep{he2021debertav3} & - \\
Nepali & nep & Generalist & mDeBERTa-v3 (Base) & - \\
Spanish & spa & Generalist & mDeBERTa-v3 (Base) & - \\
Swahili & swa & Generalist & mDeBERTa-v3 (Base) & - \\
\midrule
\multicolumn{5}{c}{\textit{Group B: Specialist Superiority}} \\
Arabic & arb & Specialist & AraBERTv02-Twitter (Large) \citep{antoun2020arabert} & - \\
Bengali & ben & Specialist & BanglaBERT (Base) \citep{bhattacharjee2022banglabert} & - \\
Chinese & zho & Specialist & MacBERT (Base) \citep{cui2020revisiting} & - \\
German & deu & Specialist & GBERT (Base) \citep{chan2020germans} & - \\
Hausa & hau & Specialist & Davlan/Hausa-XLMR \citep{alabi2022afro} & $\tau=0.35$ \\
Italian & ita & Specialist & GilBERTo \citep{gilberto2020} & - \\
Khmer & khm & Specialist & Metythorn/Khmer-XLMR (Base) \citep{metythorn2025khmer} & - \\
Odia & ori & Specialist & L3Cube-Odia \citep{joshi2023l3cubetelugu} & - \\
Telugu & tel & Specialist & L3Cube-Telugu \citep{joshi2023l3cubetelugu} & - \\
\midrule
\multicolumn{5}{c}{\textit{Group C: Hybrid Ensembles}} \\
Amharic & amh & Ensemble & Afro-XLMR \citep{alabi2022afro} + mDeBERTa-v3 & 40/60 \\
English & eng & Ensemble & DeBERTa-v3 (Large) + BERTweet \citep{nguyen2020bertweet} & 65/35, $\tau=0.45$ \\
Hindi & hin & Ensemble & mDeBERTa-v3 + L3Cube-Hindi-Hate \citep{joshi2022l3cube} & 50/50, $\tau=0.60$ \\
Persian & fas & Ensemble & ParsBERT \citep{farahani2021parsbert} + mDeBERTa-v3 & 60/40, $\tau=0.60$ \\
Polish & pol & Ensemble & XLM-R (Base) + HerBERT \citep{mroczkowski2021herbert} & 50/50, $\tau=0.45$ \\
Punjabi & pan & Ensemble & mDeBERTa-v3 + XLM-R + MuRIL \citep{khanuja2021muril} & Average \\
Russian & rus & Ensemble & DeepPavlov \citep{kuratov2019adaptation} + mDeBERTa-v3 & 50/50 \\
Turkish & tur & Ensemble & Savasy \citep{yildirim2024finetuning} + dbmdz \citep{schweter2020dbmdz} & 50/50 \\
Urdu & urd & Ensemble & MuRIL (Base) \citep{khanuja2021muril} + XLM-R (Base) & 50/50 \\
\bottomrule
\end{tabular}
\caption{Final System Configuration. Ensemble weights reflect the best-performing tried development configurations, and thresholds ($\tau$) were adjusted only where beneficial. Unless explicitly stated in the table, we used the default decision threshold of $\tau=0.5$.}
\label{tab:model_strategy}
\end{table*}

\section{Methodology}
\subsection{Phase 1: Architecture Selection Pipeline}
We established XLM-RoBERTa-Base \citep{conneau-etal-2020-unsupervised} as a universal baseline across all 22 tracks. To address the subword fragmentation that standard generalists often suffer in distinct scripts (Appendix~\ref{sec:appendix_tokenization}), we maintained a comprehensive development ledger and tested between 4 and 13 candidate models or ensemble configurations per track, with deeper search in languages where more resources were available and early candidates failed to meaningfully surpass the baseline.

This policy yielded improvements in 18 of 22 languages (e.g., Odia $+10.6\%$, Khmer $+8.1\%$). However, some tracks required tactical pivots: in \textbf{Burmese} and \textbf{Punjabi}, standalone specialists underperformed, likely due to domain mismatch between formal pretraining corpora and noisy tweets; in \textbf{Spanish}, the specialist (\textit{RoBERTuito}) yielded only a marginal $+0.6\%$ gain, prompting a pivot to mDeBERTa-v3 ($+4.7\%$); and in \textbf{Urdu}, specialists failed to separate meaningfully from the baseline, necessitating ensembles. Notably, mDeBERTa-v3's consistent cross-lingual performance also made it a core component in diverse tracks such as \textbf{Hindi} and \textbf{Persian} (Farsi).

Through this systematic evaluation of monolingual specialists, high-capacity generalists (mDeBERTa-v3), and hybrid ensembles, a non-baseline architecture was adopted only if it improved development Macro-F1 by $\geq 2\%$ or showed a more balanced Precision/Recall profile than the XLM-R baseline. In low-margin cases, this criterion was applied conservatively: small development-set differences were not treated as decisive unless accompanied by a more stable precision--recall trade-off. A representative sample of these transitions is detailed in Appendix~\ref{sec:appendix_selection}.

\subsection{Phase 2: Hybrid Ensembles \& Calibration}
For languages exhibiting complementary error profiles, we constructed weighted soft-voting ensembles to leverage complementary strengths. For instance, in \textbf{English}, we combined a domain-specific social media model (BERTweet) with a general-purpose encoder (\textit{DeBERTa-v3-Large} \citep{he2021debertav3}). The ensemble probability is calculated as $P(x) = \alpha P_{Spec}(x) + (1 - \alpha) P_{Gen}(x)$, where $\alpha \in [0,1]$ denotes the specialist weight in the soft-voting mixture (generalized to an unweighted mean for architectures with $>2$ models such as \textbf{Punjabi}).

In practice, ensemble configurations were explored as part of the broader development-phase search (Section~3.1). The reported weights therefore correspond to the best-performing tried configurations on the development set rather than to a dedicated local search over $\alpha$ alone. No additional held-out subset was created for $\alpha$ tuning; candidate ensemble variants were compared on the full development set, as the already limited validation sizes made further partitioning impractical.

We used this empirical ensemble selection strategy primarily in languages where different candidate systems exhibited complementary error profiles. Thresholds $\tau$ were additionally adjusted in a small number of cases to counteract class bias and restore more balanced decision boundaries in overly aggressive models. Despite the overfitting risks inherent to small validation samples ($N \approx 160$), such targeted calibration was often practically useful. Specifically, the standalone \textbf{Hausa} specialist and the ensembles for \textbf{Hindi}, \textbf{Persian}, \textbf{Polish}, and \textbf{English} utilized $\tau \in \{0.35, 0.60, 0.60, 0.45, 0.45\}$ respectively for precision optimization or recall support (Table~\ref{tab:model_strategy}). We did not perform a separate systematic sensitivity analysis of $\alpha$ under fixed-model $\pm 10\%$ perturbations.

\subsection{Phase 3: Cross-Lingual Augmentation}
To address RQ3, we explored cross-lingual data augmentation across 11 language tracks. Using \textbf{NLLB-200-Distilled-600M} \citep{costa2022no}, we translated the complete English training set ($\approx 3{,}200$ samples) into each target language and merged these synthetic samples with the native training data. This augmentation strategy was intended to test whether English supervision could provide useful additional signal for lower-resource tracks without requiring language-specific annotation expansion. However, as detailed in Appendix \ref{sec:appendix_translation}, the results were highly inconsistent, leading us to abandon translation for the final pipeline and prioritize native architecture tuning instead.

\section{Experimental Setup}
To maximize data, we employed a \textbf{Merge-and-Retrain} strategy: after finalizing hyperparameters during the development phase, we concatenated the Train and Dev sets. A 5\% internal split was reserved only for epoch-level evaluation logging, allowing us to monitor validation loss and basic convergence behavior during training. This split was not used for early stopping, best-checkpoint reloading, or further hyperparameter tuning; instead, each model was trained for its pre-specified number of epochs, and final predictions were generated from the resulting final model state.

All training utilized mixed precision (\texttt{fp16}), a maximum sequence length of 128, and weight decay of 0.01. Per-device batch sizes (2--16) and gradient accumulation steps were adjusted to target an effective batch size of 16 (ranging from 4 to 64). Learning rates ($5e^{-6}$ to $3e^{-5}$) and epochs (4--6) were tuned per model family, and all reported final runs used a single fixed random seed (42) per track. We did not perform a systematic multi-seed evaluation across languages; while this choice improved reproducibility and computational feasibility in a 22-language setting, it also means that seed-level variance---especially on very small development sets ($N \approx 160$)---remains a limitation. Accordingly, we interpret low-margin development differences cautiously, as some architecture selections near the decision boundary may be seed-sensitive.

\section{Results and Discussion}

\subsection{Overall Performance}
Our final results on the official test set are presented in Table \ref{tab:main_results}. The system achieved a macro-averaged F1 of \textbf{0.796} and an average accuracy of \textbf{0.826}.

For context, Appendix~\ref{app:baseline_context} compares the organizer baseline \citep{naseem-etal-2026-polar} with our XLM-R development baseline, which anchored Phase 1.

\begin{table}[h!]
\centering
\small
\setlength{\tabcolsep}{3pt}
\begin{tabular}{lrrr|lrrr}
\toprule
\textbf{Lang} & \textbf{Acc} & \textbf{F1(B)} & \textbf{F1(M)} & \textbf{Lang} & \textbf{Acc} & \textbf{F1(B)} & \textbf{F1(M)} \\
\midrule
amh & 0.824 & 0.880 & 0.774 & nep & 0.882 & 0.882 & 0.882 \\
arb & 0.832 & 0.807 & 0.829 & ori & 0.838 & 0.691 & 0.790 \\
ben & 0.847 & 0.824 & 0.845 & pan & 0.769 & 0.783 & 0.768 \\
mya & 0.889 & 0.902 & 0.887 & fas & 0.871 & 0.913 & 0.831 \\
zho & 0.891 & 0.889 & 0.891 & pol & 0.812 & 0.774 & 0.806 \\
eng & 0.810 & 0.742 & 0.796 & rus & 0.820 & 0.697 & 0.784 \\
deu & 0.711 & 0.690 & 0.710 & spa & 0.763 & 0.755 & 0.763 \\
hau & 0.931 & 0.672 & 0.817 & swa & 0.783 & 0.795 & 0.782 \\
hin & 0.896 & 0.938 & 0.803 & tel & 0.889 & 0.892 & 0.889 \\
ita & 0.630 & 0.539 & 0.615 & tur & 0.784 & 0.785 & 0.784 \\
khm & 0.927 & 0.961 & 0.711 & urd & 0.783 & 0.840 & 0.751 \\
\bottomrule
\end{tabular}
\caption{Official Test Phase Results. F1(B) = Binary-F1, F1(M) = Macro-F1, Acc = Accuracy.}
\label{tab:main_results}
\end{table}

\subsection{Leaderboard Dynamics and Performance Consistency}
To evaluate framework robustness, we compared our results against the best \textit{participant} score on the public leaderboard snapshot (SOTA), defining the performance delta as $\Delta_{SOTA} = S_{Ours} - S_{SOTA}$.\footnote{Public SemEval-2026 Task 9 leaderboard snapshot: \href{https://github.com/Polar-SemEval/Leaderboards}{https://github.com/Polar-SemEval/Leaderboards}. Here, SOTA refers to the best non-baseline participant submission in a given language.}

Given the subjectivity of polarization annotation and small development sets \citep{uma2021learning,davani2022dealing}, we interpret minor leaderboard gaps cautiously. We report a descriptive 4-point Macro-F1 window ($\Delta_{\mathrm{SOTA}} \geq -0.04$) to contextualize proximity to the public leaderboard best, rather than claiming definitive architectural superiority. Under this heuristic, our system is within 4 points of the top score on 13 tracks (Table~\ref{tab:sota_gap}), although cases near the cutoff should be interpreted cautiously.

\begin{table}[h]
\centering
\small
\begin{tabular*}{\columnwidth}{@{\extracolsep{\fill}}lcccc@{}}
\toprule
\textbf{Language} & \textbf{Rank} & \textbf{Our Score} & \textbf{SOTA} & \textbf{$\Delta_{SOTA}$} \\
\midrule
Persian (fas) & \textbf{3} & 0.8308 & 0.8348 & \textbf{-0.0040} \\
Burmese (mya) & \textbf{2} & 0.8874 & 0.8913 & \textbf{-0.0039} \\
Telugu (tel) & 7 & 0.8892 & 0.9053 & -0.0161 \\
Hausa (hau) & 6 & 0.8168 & 0.8336 & -0.0168 \\
Bengali (ben) & 7 & 0.8446 & 0.8625 & -0.0179 \\
Arabic (arb) & 12 & 0.8294 & 0.8488 & -0.0194 \\
Hindi (hin) & 17 & 0.8032 & 0.8236 & -0.0204 \\
Amharic (amh) & 10 & 0.7744 & 0.8002 & -0.0258 \\
Swahili (swa) & 15 & 0.7823 & 0.8113 & -0.0290 \\
English (eng) & 18 & 0.7958 & 0.8252 & -0.0294 \\
Odia (ori) & 14 & 0.7903 & 0.8255 & -0.0352 \\
Polish (pol) & 13 & 0.8061 & 0.8431 & -0.0370 \\
Spanish (spa) & 22 & 0.7632 & 0.8030 & -0.0398 \\
\bottomrule
\end{tabular*}
\caption{Tracks within 4 Macro-F1 points of the public leaderboard best participant submission ($\Delta_{\mathrm{SOTA}} \geq -0.04$). $\Delta_{\mathrm{SOTA}}$ denotes the raw Macro-F1 difference.}
\label{tab:sota_gap}
\end{table}

This pattern suggests relatively consistent cross-lingual competitiveness under the above descriptive heuristic, including an official \textbf{2nd-place} finish in Burmese and a shared \textbf{2nd-place} standing in Persian reported in the official task overview \citep{naseem-etal-2026-polar}, while the public leaderboard snapshot ranks the raw scores numerically.

\paragraph{Cross-Lingual Stability \& Leaderboard Density.}
Analyzing the public leaderboard snapshot reveals substantial cross-lingual variance among competing systems, which dominate in higher-resource tracks but degrade in distinct-script tracks like Khmer. By switching between generalists and specialists, our approach mitigated cross-lingual overfitting, remaining competitive in 59\% of tracks. Resource availability clearly impacts leaderboard density: higher-resource tracks (e.g., English, Spanish) are highly compressed, suggesting standard architectures have reached saturation. Conversely, in distinct-script tracks like \textbf{Persian} and \textbf{Burmese}, pivoting to a targeted ensemble and high-capacity generalist provided a measurable, highly competitive advantage. Finally, \textbf{Italian} proved uniquely challenging; a dense, low-scoring cluster ($0.60$--$0.65$) suggests distribution shifts hindered nearly all architectures, with only one isolated system reaching $0.7303$.

\subsection{Failure Modes and Error Analysis}
While our approach proved robust for the majority of languages, we observed distinct failure modes in five tracks where our system underperformed significantly relative to the current public best participant submission ($\Delta_{SOTA} < -0.05$), as detailed in Table~\ref{tab:failure_analysis}. Notably, Khmer also illustrates that a positive dev-test shift does not necessarily imply genuine generalization quality.

\begin{table}[h]
\centering
\small
\begin{tabular*}{\columnwidth}{@{\extracolsep{\fill}}lccc@{}}
\toprule
\textbf{Language} & \textbf{Macro-F1} & \textbf{SOTA} & \textbf{$\Delta_{SOTA}$} \\
\midrule
Italian (ita) & 0.6149 & 0.7303 & -0.1154 \\
German (deu) & 0.7096 & 0.7608 & -0.0512 \\
Punjabi (pan) & 0.7679 & 0.8257 & -0.0578 \\
Khmer (khm) & 0.7113 & 0.7744 & -0.0631 \\
Urdu (urd) & 0.7505 & 0.8196 & -0.0691 \\
\bottomrule
\end{tabular*}
\caption{The ``Challenge Tracks'': languages where our system underperformed by more than 5 Macro-F1 points relative to the public best participant submission ($\Delta_{SOTA} < -0.05$).}
\label{tab:failure_analysis}
\end{table}

\textbf{The Recall Trap (Khmer \& Urdu):} Both models struggled to recover the \textit{neutral} class. Khmer predicted the polarized class for $\sim$95\% of test samples, indicating majority-class collapse. Although exceeding the majority baseline, we treat this as a failure mode. We did not add remedies like focal loss or negative sampling in the final submission to maintain a consistent cross-lingual framework, as they would have required additional track-specific retraining under a limited submission budget, and such adjustments could not be reliably validated on the limited development set.

\smallskip

\textbf{The Specialist Trap (German):} Although the specialist (\textit{GBERT}) clearly outperformed the XLM-R baseline during development, its $-0.0512$ SOTA gap on the test set exposes the danger of relying on small validation splits ($N \approx 160$). This sharp reversal suggests the model likely overfit to the development data. Because GBERT showed such strong initial results with no obvious warning signs, its +6.95\% development gain ultimately proved deceptive. Furthermore, since the German track did not display the extreme class bias seen elsewhere, basic threshold calibration likely would not have resolved this fundamental failure to generalize.

\smallskip

\textbf{Semantic Overfitting (Punjabi):} In \textbf{Punjabi}, our Super-Ensemble strategy resulted in a gap of $-0.0578$. Given the massive parameter count of ensembling three models on an extremely constrained validation split ($N=100$), we hypothesize the system suffered from semantic overfitting. Rather than learning robust, generalizable decision boundaries, the combined architectures appear to have overfit to development-specific lexical cues, limiting their ability to transfer to the broader test distribution.

A further limitation is that we did not conduct a dedicated weight-sensitivity ablation around the final ensemble ratios; the reported mixtures reflect the selected development-phase configurations rather than a systematic local search over nearby $\alpha$ values.

\smallskip

\textbf{The Low-Ceiling Track (Italian):} Italian showed a different failure mode: the leaderboard was compressed in a relatively low scoring band, with most systems clustering below 0.68 and only one clear outlier at 0.7303. This suggests that the main bottleneck in Italian may lie less in architecture choice alone than in dataset-specific distributional difficulty.

\subsection{Dev-Test Shift Analysis}
To evaluate system robustness beyond absolute leaderboard position, we compared the best development-phase Macro-F1 scores against the official test results. This comparison helps distinguish genuine cross-split generalization from development-set overfitting, especially in a setting with small validation samples and substantial cross-lingual variation. We analyzed the top three gains, five largest losses, six most stable tracks, and one anomalous gain (detailed in Appendix \ref{sec:appendix_shift}).

\smallskip

\textbf{Group 1: Generalization Winners (Gains).} Telugu (+3.3\%), Burmese (+2.8\%), and Polish (+2.3\%) exhibited meaningful improvements from development to test. In all three cases, the gains were accompanied by comparatively balanced prediction ratios (41\%--56\%), suggesting that the selected architectures generalized with robust decision boundaries rather than benefiting from trivial label bias.

\smallskip

\textbf{Special Case: Anomalous Positive Delta (Khmer).} Although Khmer recorded the largest numerical dev-test gain ($+4.1\%$), we do not interpret this as a genuine generalization success. The model predicted the polarized class for 95.6\% of test samples, indicating majority-class collapse rather than robust semantic discrimination. We therefore analyze Khmer as a pathological gain case: mathematically, it belongs among the largest positive deltas, but behaviorally it is better understood alongside the Challenge Tracks in Table \ref{tab:failure_analysis}.

The Khmer test set is highly skewed (90.8\% polarized). While our 0.7113 score meaningfully exceeds a trivial majority-class baseline (0.4758), the extreme prediction skew confirms weak recovery of the neutral class rather than robust semantic generalization.

\smallskip

\textbf{Group 2: Stable Performers.} Spanish, Arabic, Bengali, Nepali, Amharic, and Chinese showed remarkable stability ($|\Delta| \leq 2.0\%$). For Arabic, Bengali, and Chinese, high-quality monolingual pretraining (e.g., AraBERT) acted as a strong regularizer. For Nepali, stability (+1.2\%) highlighted the sample efficiency of mDeBERTa-v3's replaced token detection objective. Finally, Amharic (-0.9\%) demonstrated that weighted ensembling effectively mitigates individual model variance.

\smallskip

\textbf{Group 3: Overfitting Regressions (Losses).} The largest drops occurred in Hindi ($-8.5\%$), Italian ($-7.8\%$), Urdu ($-7.7\%$), Persian ($-6.6\%$), and Punjabi ($-6.2\%$). In Italian and Punjabi, models maintained class balance but performance collapsed, suggesting \textit{semantic overfitting} to dev-set lexical artifacts. Conversely, Hindi, Persian, and Urdu suffered from excessive positive predictions ($67\%$--$83\%$), indicating that while threshold calibration ($\tau \geq 0.60$) was necessary, it was insufficient to fully counteract inherent ``trigger-happy'' biases on these test sets.

Taken together, these patterns suggest that multilingual polarization detection is shaped less by model size alone than by the interaction among tokenizer fit, domain alignment, and calibration under limited supervision.

\subsection{Discussion of Research Questions}
\textbf{RQ1: Does XLM-RoBERTa provide a sufficient universal baseline across 22 diverse languages? No.} While XLM-RoBERTa offers a convenient starting point, it struggles significantly with distinct or underrepresented scripts. As established in Phase 1, monolingual specialists outperformed the initial XLM-RoBERTa baseline in the vast majority of our 22 tracks (e.g., Odia $+10.6\%$, Khmer $+8.1\%$). Furthermore, our fragmentation analysis (Appendix \ref{sec:appendix_tokenization}) suggests XLM-R fragments text up to 38\% more than native specialists. This lexical bottleneck suggests that the architectural capacity of a standard generalist often cannot compensate for a lack of specialized vocabulary. Consequently, achieving competitive performance required abandoning the universal baseline in favor of culturally grounded monolingual specialists or alternative high-capacity generalists (e.g., mDeBERTa-v3).

\smallskip

\textbf{RQ2: Under what linguistic conditions do specialists or alternative generalists outperform the baseline?} Our findings suggest specialists excel under two primary conditions: tokenizer efficiency and domain alignment. In \textbf{Arabic} (\textit{AraBERT}) and \textbf{Khmer}, specialists resolved subword fragmentation bottlenecks \textit{and} aligned with informal social media text. However, localized vocabulary alone is insufficient. In \textbf{Burmese} and \textbf{Nepali}, native specialists trained on formal literature suffered severe domain mismatch on tweets, losing to mDeBERTa-v3. Furthermore, high-capacity generalists occasionally outperformed even domain-adapted specialists (e.g., \textit{RoBERTuito} in \textbf{Spanish}), indicating that architectural depth can sometimes eclipse localized pretraining.

\smallskip

\textbf{RQ3: Can cross-lingual data augmentation compensate for resource scarcity?} In our specific pipeline, exploratory cross-lingual transfer via NLLB-200 yielded mixed results (Appendix \ref{sec:appendix_translation}). While translating the $\approx 3,200$ English source samples marginally lifted weak XLM-R baselines in \textbf{German, Spanish, and Turkish}, these minor gains were ultimately superseded by simply selecting stronger unaugmented architectures. Moreover, morphologically rich languages suffered severe degradation (\textbf{Russian} dropped $0.743 \to 0.684$, \textbf{Polish} $0.760 \to 0.660$), suggesting the translation model struggled to generate the precise inflections required by native tokenizers. While more advanced, large-scale augmentation paradigms may succeed, this naive forward translation proved too unstable for our pipeline, leading us to prioritize native architecture selection.

\section{Conclusion}
In this study, we addressed multilingual polarization detection across 22 languages through a language-adaptive selection strategy. Our results show that this task resists a one-size-fits-all solution: while XLM-RoBERTa provides a useful universal starting point, strong performance often required language-specific specialists, alternative high-capacity generalists such as mDeBERTa-v3, or weighted hybrid ensembles with targeted threshold calibration. Cross-lingual translation-based augmentation was comparatively unstable, especially in morphologically rich languages. Overall, the findings support language-specific architecture selection as a more reliable strategy than universal model scaling alone or unfiltered synthetic augmentation. This highlights the ongoing necessity for culturally and linguistically grounded NLP solutions.

\section*{Limitations}
Our study has several methodological and computational limitations. First, architecture selection and threshold calibration relied on small validation splits ($N \approx 160$), so some decisions may be split- or seed-sensitive, and we did not perform multi-seed evaluation. Second, due to submission and computational constraints, we did not conduct local ensemble-weight sensitivity analyses or apply class-imbalance remedies (e.g., focal loss, negative sampling) for heavily skewed tracks such as Khmer or Urdu. Finally, our augmentation experiments relied on a 600M-parameter distilled NLLB-200 model in a simple forward-translation setup. The degradation observed in morphologically rich languages should therefore be interpreted as a limitation of this pipeline, not as a general rejection of multilingual data augmentation.

\section*{Acknowledgments}
I thank the organizers of SemEval-2026 Task 9. I dedicate the effort and late nights behind this work to my people in Iran, whose resilience and courage continue to inspire me every day.

\bibliography{custom}

\appendix
\section{Appendix}

\subsection{Tokenization Fragmentation Analysis}
\label{sec:appendix_tokenization}
The \textit{Fragmentation Ratio} (average subwords per word) measures tokenizer efficiency. High fragmentation wastes attention on lexical reconstruction over semantic modeling. Table \ref{tab:tokenization} suggests that native specialists reduce fragmentation by up to 38.0\% for non-Latin and morphologically rich languages.

\begin{table}[h]
\centering
\small
\begin{tabular}{llcc}
\toprule
\textbf{Lang} & \textbf{XLM-R} & \textbf{Specialist} & \textbf{Reduction} \\
\midrule
German & 1.60 & 1.46 (\textit{GBERT}) & 8.8\% \\
Polish & 1.90 & 1.58 (\textit{HerBERT}) & 16.8\% \\
Persian & 1.53 & 1.17 (\textit{ParsBERT}) & 23.5\% \\
Arabic & 1.85 & 1.37 (\textit{AraBERT}) & 25.9\% \\
Bengali & 1.84 & 1.14 (\textit{BanglaBERT}) & 38.0\% \\
\bottomrule
\end{tabular}
\caption{Fragmentation Ratio (lower is better). This subset represents complex morphology (German, Polish) and distinct scripts (Persian, Arabic, Bengali).}
\label{tab:tokenization}
\end{table}

\subsection{Development Architecture Selection}
\label{sec:appendix_selection}
As detailed in Section~3.1, our strategy required a $\geq 2\%$ Macro-F1 improvement to justify pivoting from the XLM-R baseline to an alternative architecture. Table~\ref{tab:dev_selection} provides a representative snapshot of this development ledger, illustrating where specialist models, stronger generalists, or ensembles cleared that selection threshold.

\begin{table}[ht]
\centering
\small
\setlength{\tabcolsep}{3pt}
\resizebox{\columnwidth}{!}{
\begin{tabular}{llccc}
\toprule
\textbf{Lang} & \textbf{Selected Architecture} & \textbf{XLM-R} & \textbf{Selected} & \textbf{$\Delta_{Dev}$} \\
\midrule
\multicolumn{5}{c}{\textit{Baseline $\to$ Monolingual Specialist}} \\
\midrule
Odia & L3Cube-Odia & 0.7257 & 0.8317 & \textbf{+10.60\%} \\
Khmer & Metythorn/Khmer-XLMR & 0.5888 & 0.6696 & \textbf{+8.08\%} \\
German & GBERT (Base) & 0.6700 & 0.7395 & \textbf{+6.95\%} \\
Hausa & Davlan/Hausa & 0.7834 & 0.8488 & \textbf{+6.54\%} \\
Arabic & AraBERTv02 (Large) & 0.7972 & 0.8272 & \textbf{+3.00\%} \\
\midrule
\multicolumn{5}{c}{\textit{Baseline $\to$ High-Capacity Generalist}} \\
\midrule
Spanish & mDeBERTa-v3 (Base) & 0.6964 & 0.7439 & \textbf{+4.75\%} \\
Nepali & mDeBERTa-v3 (Base) & 0.8500 & 0.8700 & \textbf{+2.00\%} \\
\midrule
\multicolumn{5}{c}{\textit{Baseline $\to$ Hybrid Ensemble}} \\
\midrule
Hindi & mDeBERTa-v3 + L3Cube-Hindi & 0.7750 & 0.8882 & \textbf{+11.32\%} \\
Turkish & Super-Ensemble & 0.7390 & 0.8075 & \textbf{+6.85\%} \\
Persian & ParsBERT + mDeBERTa-v3 & 0.8150 & 0.8969 & \textbf{+8.19\%} \\
Amharic & Afro-XLMR + mDeBERTa-v3 & 0.7162 & 0.7831 & \textbf{+6.69\%} \\
Punjabi & XLM-R + MuRIL + mDeBERTa-v3 & 0.7799 & 0.8296 & \textbf{+4.97\%} \\
\bottomrule
\end{tabular}
}
\caption{Architecture Selection: Development Macro-F1 comparisons between the initial XLM-R baseline and the adopted architecture.}
\label{tab:dev_selection}
\end{table}

\subsection{Translation Ablation Results}
\label{sec:appendix_translation}
Table \ref{tab:translation_ablation} compares the XLM-R baseline, the translation-augmented variant, and our final submitted architectures. It provides a compact view of whether synthetic translated data offered meaningful gains over the baseline and how those results compared with the stronger architectures ultimately selected for submission.

\begin{table}[ht]
\centering
\small
\setlength{\tabcolsep}{3pt}
\resizebox{\columnwidth}{!}{
\begin{tabular}{llccc}
\toprule
\textbf{Lang} & \textbf{Aug. Model} & \textbf{Baseline} & \textbf{Augmented} & \textbf{Final} \\
\midrule
Arabic & MARBERT \citep{abdul-mageed-etal-2021-arbert} & 0.797 & 0.797 & \textbf{0.829} \\
German & GBERT-Base & 0.670 & 0.699 & \textbf{0.710} \\
Italian & UmBERTo \citep{parisi2021umberto} & 0.646 & 0.613 & \textbf{0.615} \\
Odia & MuRIL-Base & 0.726 & 0.731 & \textbf{0.790} \\
Punjabi & MuRIL-Base & 0.780 & 0.700 & \textbf{0.768} \\
Polish & PolBERT \citep{Kleczek2020} & 0.760 & 0.660 & \textbf{0.806} \\
Russian & RuBERT \citep{rubert-2024} & 0.743 & 0.684 & \textbf{0.784} \\
Spanish & BETO \citep{canete2020spanish} & 0.696 & 0.702 & \textbf{0.763} \\
Swahili & Afro-XLMR & 0.779 & \textbf{0.791} & 0.782 \\
Turkish & dbmdz & 0.739 & 0.756 & \textbf{0.784} \\
Urdu & MuRIL-Base & 0.722 & 0.705 & \textbf{0.751} \\
\bottomrule
\end{tabular}
}
\caption{Translation Ablation (Macro-F1). Augmentation destabilized highly inflected languages and underperformed our final architectures in 10 of 11 tracks.
}
\label{tab:translation_ablation}
\end{table}

\subsection{Organizer Baseline vs. XLM-R Baseline}
\label{app:baseline_context}
Table~\ref{tab:baseline_context} compares the official Subtask~1 LaBSE baseline \citep{naseem-etal-2026-polar} with our in-house XLM-R dev baseline, which served as the reference point for our early architecture decisions.

\begin{table}[ht]
\centering
\small
\begin{tabular*}{\columnwidth}{@{\extracolsep{\fill}}lcc@{}}
\toprule
\textbf{Language} & \textbf{Organizer Baseline} & \textbf{XLM-R (Dev)} \\
\midrule
Amharic (amh) & 0.764 & 0.716 \\
Arabic (arb) & 0.812 & 0.797 \\
Bengali (ben) & 0.825 & 0.839 \\
Burmese (mya) & 0.861 & 0.857 \\
Chinese (zho) & 0.864 & 0.888 \\
English (eng) & 0.773 & 0.784 \\
German (deu) & 0.686 & 0.670 \\
Hausa (hau) & 0.821 & 0.783 \\
Hindi (hin) & 0.782 & 0.775 \\
Italian (ita) & 0.564 & 0.646 \\
Khmer (khm) & 0.737 & 0.588 \\
Nepali (nep) & 0.883 & 0.850 \\
Odia (ori) & 0.776 & 0.725 \\
Punjabi (pan) & 0.749 & 0.779 \\
Persian (fas) & 0.835 & 0.815 \\
Polish (pol) & 0.773 & 0.759 \\
Russian (rus) & 0.748 & 0.742 \\
Spanish (spa) & 0.750 & 0.696 \\
Swahili (swa) & 0.790 & 0.779 \\
Telugu (tel) & 0.889 & 0.855 \\
Turkish (tur) & 0.750 & 0.739 \\
Urdu (urd) & 0.742 & 0.722 \\
\bottomrule
\end{tabular*}
\caption{Comparison of the organizer-provided baseline and our in-house XLM-R development baseline.}\label{tab:baseline_context}
\end{table}

\subsection{Dev-Test Shift Analysis}
\label{sec:appendix_shift}
Table~\ref{tab:dev_test_shift} reports the best Macro-F1 delta between development and test. To evaluate system robustness, the 15 analyzed tracks are grouped to illustrate the effects of distribution shift, architectural robustness, and threshold calibration.

\begin{table}[H]
\centering
\small
\begin{tabular*}{\columnwidth}{@{\extracolsep{\fill}}lccc@{}}
\toprule
\textbf{Language} & \textbf{Dev F1} & \textbf{Test F1} & \textbf{$\Delta$} \\
\midrule
\multicolumn{4}{c}{\textit{Special Case: Anomalous Gain}} \\
\midrule
Khmer (khm) & 0.670 & 0.711 & \textbf{+4.1\%} \\
\midrule
\multicolumn{4}{c}{\textit{Group 1: Top Gains}} \\
\midrule
Telugu (tel) & 0.856 & 0.889 & \textbf{+3.3\%} \\
Burmese (mya) & 0.859 & 0.887 & \textbf{+2.8\%} \\
Polish (pol) & 0.783 & 0.806 & \textbf{+2.3\%} \\
\midrule
\multicolumn{4}{c}{\textit{Group 2: Stable Performers}} \\
\midrule
Spanish (spa) & 0.744 & 0.763 & +1.9\% \\
Nepali (nep) & 0.870 & 0.882 & +1.2\% \\
Arabic (arb) & 0.827 & 0.829 & +0.2\% \\
Bengali (ben) & 0.846 & 0.845 & -0.1\% \\
Amharic (amh) & 0.783 & 0.774 & -0.9\% \\
Chinese (zho) & 0.911 & 0.891 & -2.0\% \\
\midrule
\multicolumn{4}{c}{\textit{Group 3: Top Losses}} \\
\midrule
Punjabi (pan) & 0.830 & 0.768 & \textbf{-6.2\%} \\
Persian (fas) & 0.897 & 0.831 & \textbf{-6.6\%} \\
Urdu (urd) & 0.828 & 0.751 & \textbf{-7.7\%} \\
Italian (ita) & 0.693 & 0.615 & \textbf{-7.8\%} \\
Hindi (hin) & 0.888 & 0.803 & \textbf{-8.5\%} \\
\bottomrule
\end{tabular*}
\caption{Performance Shift: Best Development Phase Macro-F1 vs. Official Test Phase Macro-F1.}
\label{tab:dev_test_shift}
\end{table}

\end{document}